\documentclass[sigconf]{acmart}
\AtBeginDocument{%
  }

\copyrightyear{2025}
\acmYear{2025}
\setcopyright{acmlicensed}
\acmConference[MM '25] {Proceedings of the 33rd ACM International Conference on Multimedia}{October 27--31, 2025}{Dublin, Ireland.}
\acmBooktitle{Proceedings of the 33rd ACM International Conference on Multimedia (MM '25), October 27--31, 2025, Dublin, Ireland}
\acmISBN{979-8-4007-2035-2/2025/10}
\acmDOI{10.1145/3746027.3755582}

\usepackage{tabularx}
\usepackage{multirow}
\usepackage{booktabs}
\usepackage{pifont}
\usepackage{algorithm}
\begin{document}

\title{Spatial Imputation Drives Cross-Domain Alignment for EEG Classification}

\author{Hongjun Liu}
\orcid{0009-0002-2661-8047}
\affiliation{%
  \department{School of Intelligence Science and Technology} 
  \institution{University of Science and Technology Beijing}
  \city{Beijing}
  \country{China}
}
\email{D202210386@xs.ustb.edu.cn}

\author{Chao Yao}
\orcid{0000-0001-5483-3225}
\authornote{Corresponding authors}
\affiliation{%
  \department{School of Computer and Communication Engineering} 
  \institution{University of Science and Technology Beijing}
  \city{Beijing}
  \country{China}
}
\email{yaochao@ustb.edu.cn}

\author{Yalan Zhang}
\orcid{0000-0002-8736-7125}
\affiliation{%
  \department{School of Intelligence Science and Technology} 
  \institution{University of Science and Technology Beijing}
  \city{Beijing}
  \country{China}
}
\email{zhangyl@ustb.edu.cn}

\author{Xiaokun Wang}
\orcid{0009-0006-8984-9212}
\affiliation{%
  \department{School of Intelligence Science and Technology} 
  \institution{University of Science and Technology Beijing}
  \city{Beijing}
  \country{China}
}
\email{wangxiaokun@ustb.edu.cn}

\author{Xiaojuan Ban}
\orcid{0000-0001-9142-3276}
\authornotemark[1]
\affiliation{%
  \department{School of Intelligence Science and Technology} 
  \institution{University of Science and Technology Beijing}
  \city{Beijing}
  \country{China}
}
\email{banxj@ustb.edu.cn}

\renewcommand{\shortauthors}{Hongjun Liu, Chao Yao, Yalan Zhang, Xiaokun wang, \& Xiaojuan Ban}

\begin{abstract}
Electroencephalogram (EEG) signal classification faces significant challenges due to data distribution shifts caused by heterogeneous electrode configurations, acquisition protocols, and hardware discrepancies across domains. 
This paper introduces IMAC, a novel channel-dependent mask and imputation self-supervised framework that formulates the alignment of cross-domain EEG data shifts as a spatial time series imputation task.
To address heterogeneous electrode configurations in cross-domain scenarios, IMAC first standardizes different electrode layouts using a 3D-to-2D positional unification mapping strategy, establishing unified spatial representations. 
Unlike previous mask-based self-supervised representation learning methods, IMAC introduces spatio-temporal signal alignment. This involves constructing a channel-dependent mask and reconstruction task framed  as a low-to-high resolution EEG spatial imputation problem. Consequently, this approach simulates cross-domain variations such as channel omissions and temporal instabilities, thus enabling the model to leverage the proposed imputer for robust signal alignment during inference.
Furthermore, IMAC incorporates a disentangled structure that separately models the temporal and spatial information of the EEG signals separately, reducing computational complexity while enhancing flexibility and adaptability. 
Comprehensive evaluations across $10$ publicly available EEG datasets demonstrate IMAC's superior performance, achieving state-of-the-art classification accuracy in both cross-subject and cross-center validation scenarios. Notably, IMAC shows strong robustness under both simulated and real-world distribution shifts, surpassing baseline methods by up to $35$\% in integrity scores while maintaining consistent classification accuracy.
\end{abstract}

\ccsdesc[500]{Computing methodologies~Neural networks}
\ccsdesc[500]{Human-centered computing~Ubiquitous and mobile computing systems and tools}

\keywords{EEG Signal Imputation; EEG Classification; Brain-Computer Interface; Domain Adaptation}


\maketitle

\section{Introduction}

\begin{figure}
  \centering
  \includegraphics[width=\columnwidth, trim=0 20 0 10, clip]{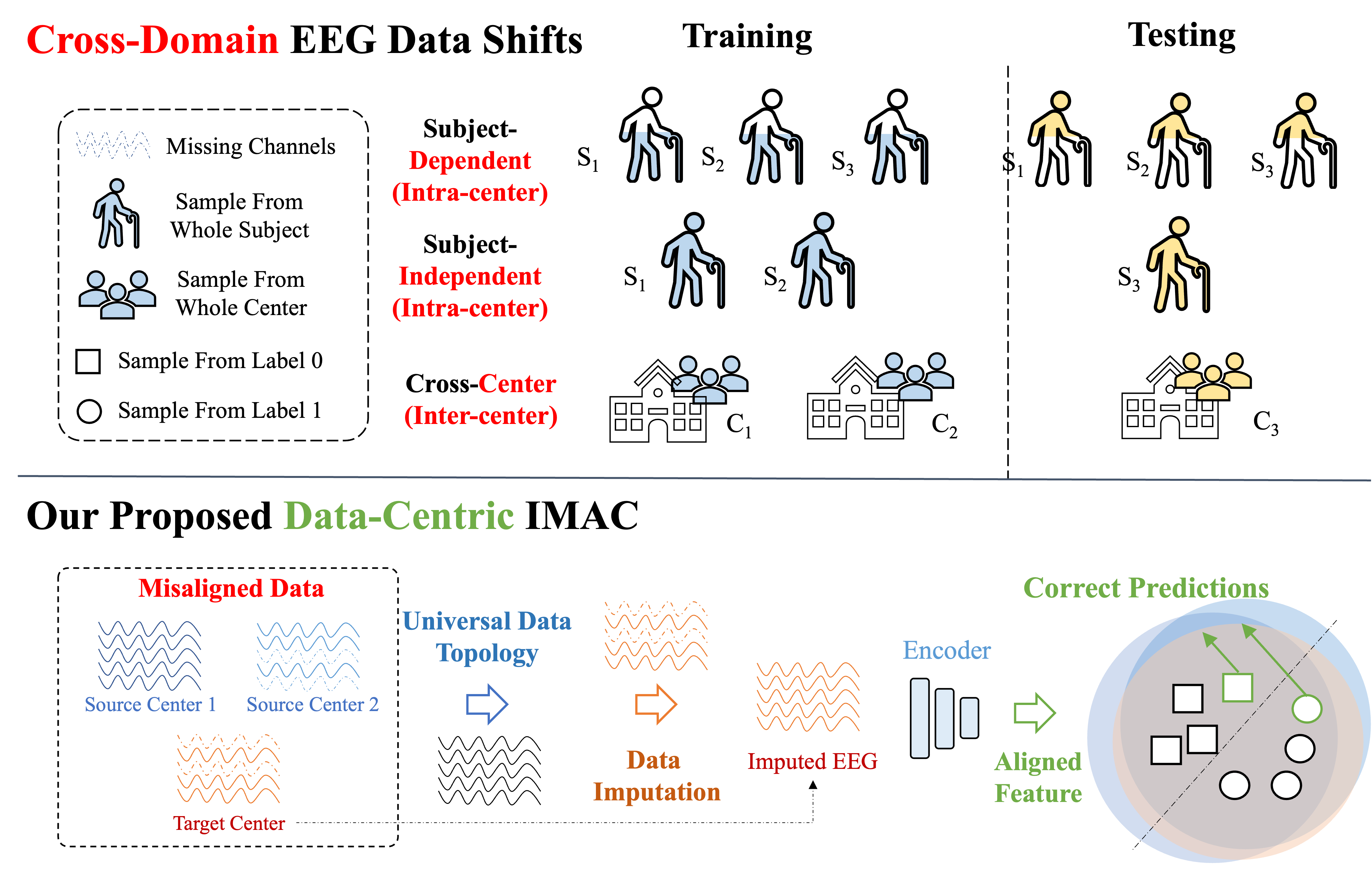}
  \caption{ Cross-domain data shifts in real-world setups. In the subject-dependent setup, samples from the same subject can appear in both the training and test sets, causing information leakage. In a subject-independent setup, samples from the same subject are exclusively in either the training or test set, which is more challenging and practically meaningful but less studied. The proposed IMAC aims to perform a cross-domain signal alignment via spatial mask and imputation.}
  \label{fig:comparison}
\end{figure}

Noninvasive electroencephalography (EEG) is a type of multichannel signal that reflects neural activity and latent psychological states, collected from electrodes deployed on the human scalp. Effective classification of EEG signals is crucial for the development of brain-computer interfaces (BCI) with applications in rehabilitation \cite{jianglarge}, emotion recognition \cite{9762054}, motor imagery \cite{HUANG2023109838}, and the diagnosis and monitoring of neurological disorders \cite{wang2024medformer}.
However, EEG classifiers often fail in real-world deployment settings due to dataset or distribution shifts, including variations in sensor location configuration, non-uniform electrode distribution, and the dynamic nature of the data \cite{10.1093/nsr/nwaf086}\cite{ragab2023source}. As shown in Figure~\ref{fig:comparison}, these shifts are common causes of domain gaps in both cross-center and cross-subject scenarios \cite{wagh2022evaluating}. Consequently, achieving consistent classification performance across both cross-subject and cross-center EEG data shifts remains a primary challenge for EEG classification.

Traditionally, numerous studies have explored transfer learning to address domain gaps, with the assumption that a large amount of unlabeled data or a small amount of labeled data from the target subjects is available \cite{Sun2025PR}. These methods generally fall into two categories: the first compares source and target domain distributions simultaneously, enabling the model to learn domain-invariant features \cite{she2023improved}\cite{9804766}; the second uses a source-free mode, where the model is pre-trained on source domain data and then fine-tuned on a large amount of unlabeled data or a small amount of labeled data from the target domain \cite{ragab2023source}\cite{xu2023dual}. 

However, when the target subjects are entirely unseen during training, the model must be trained solely on source domain data to create a domain-invariant model. Under this scenario, domain generalization (DG) approaches aim to enable the model to generalize across unseen target domains without requiring additional training data \cite{9782500}. These approaches can be broadly categorized into feature-based methods and metric-based methods. Feature-based methods, such as data augmentation, domain adversarial training, and domain alignment, focus on learning domain-invariant features. For instance, data augmentation generates synthetic samples to simulate potential domain shifts \cite{wang2024dmmr}, while domain adversarial training encourages the model to extract features that are indistinguishable across domains \cite{jia2021multi}.
Metric-based methods, in contrast, learn a domain-agnostic metric space where data from different domains are mapped into similar representations \cite{10198467}. Techniques like prototypical networks aim to align domains by learning this shared metric space \cite{qiu2024novel}. 

While these methods address domain shifts by training models to learn shared features across domains, they encounter significant limitations under conditions of pronounced heterogeneity \cite{liu2023temporal}\cite{wang2023prompt}. These limitations stem from their primary focus on learning domain-invariant features or aligning data distributions across domains without directly addressing the underlying causes of signal discrepancies. In contrast, modifying or correcting the data itself between domains would resolve the domain adaptation problem, rather than merely mitigating their effects through feature learning \cite{10.1093/nsr/nwaf086}.

To address the data shift challenges specific to EEG classification, in this paper, we tackle data distribution misalignment through spatial time series imputation. We introduce a universal imputation framework called IMpute And Classify (IMAC). The core innovation of IMAC lies in its channel masking and spatial imputation self-supervised learning strategy, which simulates diverse real-world data shifts to effectively bridge cross-domain discrepancies.
Specifically, we begin by employing spatial topology unification to standardize the electrode configuration projected into a unified 2D spatial representation.
We then apply random channel masking and perform imputation using contextual information, enabling the model to learn cross-domain correlations through a combination of consistency and fidelity losses. This process simulates cross-domain variations and promotes alignment in both spatial and temporal dimensions.
Furthermore, we propose a temporal pattern decomposition strategy, which exploits temporal structures from multiple source domains to reduce the complexity of the spatial imputation task while improving computational efficiency.
Experiments conducted across $10$ public EEG datasets demonstrate that under both simulated and real-world domain shifts, IMAC surpasses baseline methods by up to $35$\% in integrity scores while preserving consistent EEG classification accuracy, validating its ability to maintain performance stability under heterogeneous recording conditions.
The primary contributions of this work can be summarized as follows.
\begin{itemize}
    \item We formulate the issue of real-world data shifts as a spatial imputation problem. To the best of our knowledge, this is the first work to address cross-domain discrepancies by focusing on signal alignment rather than traditional feature alignment approaches.
    \item We propose a Temporal-Spatial Decomposition Module thatdisentangles EEG signals into independent temporal and spatial components. By incorporating a temporal component matching strategy, the model adaptively captures relevant temporal patterns while learning spatial correlations across channels.
    \item We propose a Channel-dependent Mask and Imputation Module that employs adaptive channel masking to simulate cross-domain variations, leveraging a self-supervised learning strategy to infer complete spatial relationships and reconstruct spatio-temporal signal omissions through a hybrid loss combining consistency and fidelity constraints.
\end{itemize}

\section{Related Works}

\subsection{Domain Generalization} 
Domain Generalization aims to improve the model’s generalization performance on inaccessible test domains utilizing a series of source domains. The current DG methods can be generally divided into three categories: 1) Domain-invariant feature learning, i.e., learning invariant representations on source domains by adding regularization terms. The author of \cite{jia2021multi} introduced a multi-view spatial-temporal graph convolutional network (MSTGCN) with domain generalization to address challenges in sleep stage classification by utilizing spatial-temporal features from multi-channel brain signals and improving model generalization across subjects. 2) Data manipulation. The author of \cite{10198467} proposed a novel domain-generalized EEG classification framework named FDCL to improve the generalizability of EEG decoding methods in unseen subjects by integrating feature decorrelation, data augmentation, and consistency learning regularizations.
Study \cite{wang2024dmmr} proposed the Denoising Mixed Mutual Reconstruction (DMMR) model to enhance the generalization of EEG-based emotion analysis across subjects through a two-stage pre-training and fine-tuning approach, incorporating self-supervised learning, data augmentation, and noise injection methods.  3) Learning strategies. The author of \cite{qiu2024novel} propose a novel multiscale convolutional prototype network (MCPNet) for automated Parkinson’s disease detection using EEG, integrating multiscale CNN and prototype learning to enhance feature diversity and improve generalization performance. In this work, we introduce a novel data manipulation technique that employs spatial masking and interpolation to achieve direct signal-level alignment, thereby fundamentally reducing distributional discrepancies.

\subsection{EEG Spatial Imputation} 
Super-resolution, initially introduced in image processing, has been widely applied to tasks involving resolution reconstruction in images and videos \cite{Jin2023ACM}\cite{liu2023JNMR}\cite{tang2024semantic}\cite{tang2024seeclear}. 
Inspired by these advancements, super-resolution techniques have also been adopted and developed for EEG signal processing to address the high cost of high-density EEG devices, named EEG spatial Imputation. 
Researchers \cite{petrichella2016channel} and \cite{courellis2016eeg} proposed interpolation-based methods to generate data for additional EEG signal channels.
To address the limited effectiveness in multichannel scenarios, the author of \cite{9796118} introduced a deep EEG super-resolution framework called Deep-EEGSR, which employs sample-specific dynamic convolutions to adapt filter parameters based on individual functional connectivity patterns. 
Building on this concept, the author of \cite{wang2025generative} proposed a spatio-temporal adaptive diffusion model (STAD) that enhances the spatial resolution of low-density EEG data using diffusion models and multi-scale Transformer denoising modules. To address the issue of EEG channel selection methods establish a unified paradigm for EEG acquisition devices, ESTformer \cite{li2023estformer} propose a multihead self-attention mechanism to the space and time dimensions, which can learn spatial structural correlations and temporal functional variations.
However, the aforementioned methods did not address three critical issues: real-world EEG variability may arise due to diverse physiological states during acquisition, recording conditions and protocols, and internals of the EEG hardware used.

\section{Methods}

\begin{figure*}[]
  \centering
  \includegraphics[width=2.0\columnwidth, trim=0 8 0 0, clip]{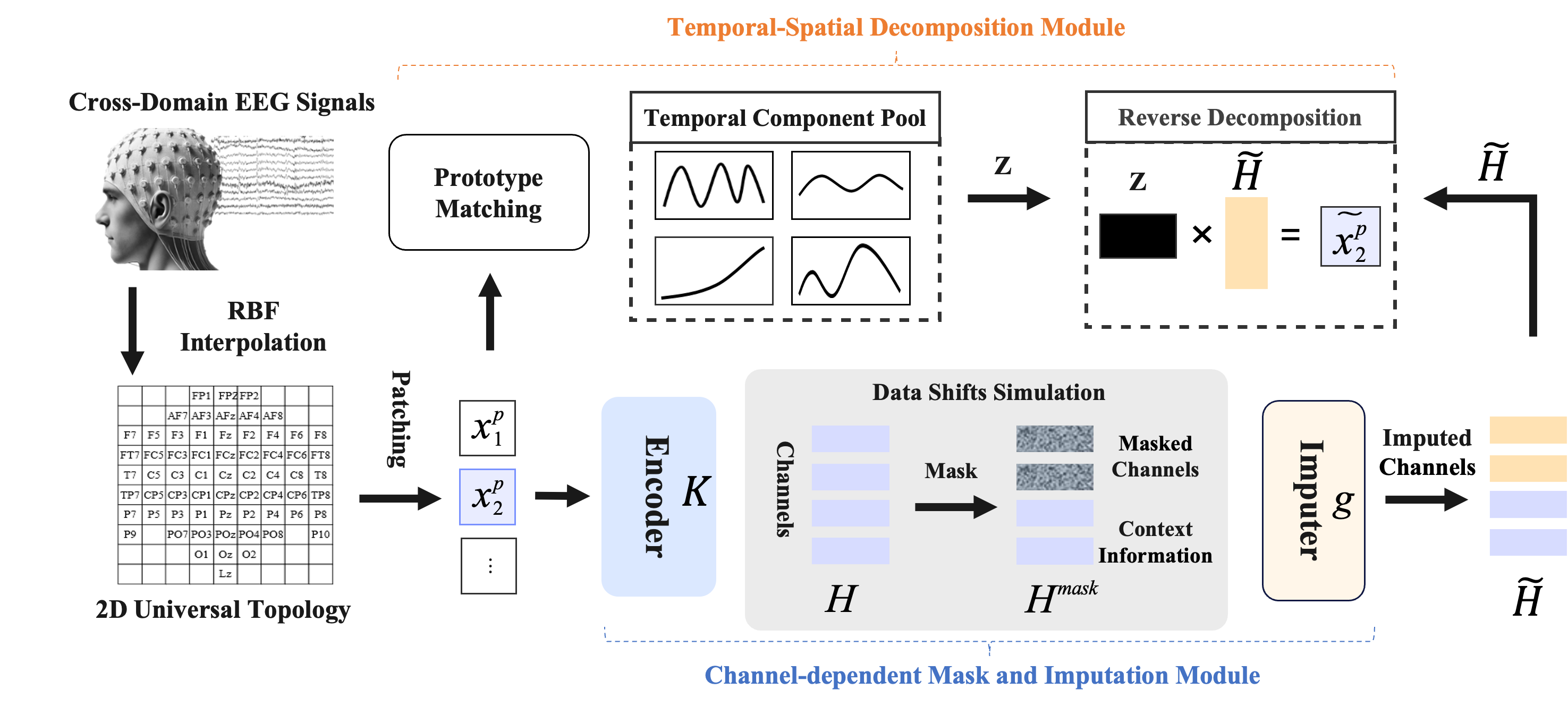}
  \caption{The training procedure of IMAC. The cross-domain EEG signals are first processed through spatial topology unification via RBF Interpolation. These preprocessed patches are then passed through an Encoder $f$ for feature extraction. The extracted features $H$ are randomly masked along the channel dimension and subsequently reconstructed by the imputer module. Simultaneously, each patch of the input signals is decoupled with temporal pattern selection, which is then multiplied with the extracted features to reconstruct the initial EEG data. }
  \label{fig: illustration}
\end{figure*}

The overall framework of the proposed method is illustrated in Figure~\ref{fig: illustration}. For cross-domain alignment, we begin by employing spatial topology unification to standardize the electrode configurations. With the EEG signals standardized, the proposed Temporal-Spatial Decomposition Module (TSDM) isolates the temporal dynamics and models the channel information separately, facilitating spatial imputation. Lastly, the Channel-dependent Mask and Imputation Module (CMIM) is designed to enable cross-domain knowledge transfer and achieve spatial EEG signal alignment effectively.

\subsection{Spatial Topology Unification Module}

The proposed Spatial Topology Unification Module (STUM) addresses the challenge of variable channel selection by unifying the spatial topology of EEG signals across multiple datasets. To achieve this, we first define the electrode coordinates based on the widely adopted 10-20 system \cite{jasper1958ten}, which provides standardized electrode positions on the scalp. Specifically, the normalized coordinates \( x_{\text{norm}} \) and \( y_{\text{norm}} \) for each electrode are scaled and rounded to the nearest integer, thereby mapping them onto a \( 9 \times 10 \) matrix. In this matrix, each electrode is assigned a position corresponding to its relative location on the scalp. The matrix is then populated with the EEG signal values, offering a 2D spatial representation of the electrode distribution. However, due to variations in channel configurations across datasets, some electrodes may be missing in certain setups. To address this issue, we apply radial basis function (RBF) interpolation to estimate the missing channels. As a result, we obtain a unified 64-channel spatial topology, ensuring consistency across all datasets.

Following spatial interpolation, we perform cross-channel patching, treating each patch as the unit for subsequent decomposition and spatial imputation. Given an input EEG sample \( x_{\text{in}} \in \mathbb{R}^{T \times C} \), we partition it into \( N \) non-overlapping cross-channel patches \( x^{(p)} = \{x^{(p)}_1, ..., x^{(p)}_N\} \), where each patch \( x^{(p)}_i \in \mathbb{R}^{N \times (L \cdot C)} \), with \( T \) representing the number of time steps, \( C \) the number of channels, and \( N \) the number of patches needed to cover the entire input signal. Finally, for each patch \( x^{(p)}_i \), we add a positional embedding feature corresponding to its position, resulting in the set of EEG patches \( \hat{x}^{(p)} = \{\hat{x}^{(p)}_1, ..., \hat{x}^{(p)}_l\} \).

\subsection{Temporal-Spatial Decomposition Module}

To tackle the challenge of spatio-temporal imputation in EEG data, our IMAC framework aims to recover complete spatial relationships across EEG channels. However, due to the strong coupling between temporal dynamics and spatial distributions in EEG signals, direct spatial reconstruction is particularly difficult \cite{toller2019sazed}.

To address this, we propose a  Temporal-Spatial Decomposition Module that decouples EEG signals into separate temporal and spatial components, allowing the model to learn temporal dynamics and spatial structures independently. In this module,  temporal patterns  are encoded as  learnable global embeddings  \( Z \), while  spatial correlations  are captured by a  channel-dependent matrix  \( H \in \mathbb{R}^{n \times d} \), where \( n \) is the number of channels and \( d \) is the embedding dimension.

Given the non-stationary nature of EEG signals, which often exhibit distributional shifts across time and domains, we design a  temporal pattern pool  that contains pre-learned embeddings representing three fundamental temporal components:  trend ,  seasonality , and  residuals . Each component is represented as a fixed embedding \( Z^T \), \( Z^S \), and \( Z^R \in \mathbb{R}^{D \times L} \), where \( D \) is the temporal feature dimension and \( L \) is the length of the time series segment. These embeddings act as reusable bases that encode different types of temporal behavior, enabling adaptive temporal modeling.

To dynamically select the most relevant temporal pattern for a given input patch \( \hat{x}^{(p)} \), we introduce a  temporal selection function  \( \mathcal{S}(\cdot) \), which computes the cosine similarity between the input and each predefined temporal component. The selected component is the one with the highest similarity score:
\begin{equation}
\mathcal{S}(\hat{x}^{(p)}) = \arg \max_{k \in \{T, S, R\}} \frac{\hat{x}^{(p)} \cdot Z^k}{\|\hat{x}^{(p)}\| \|Z^k\|}.
\end{equation}

This selection process ensures that the model utilizes the most appropriate temporal representation based on the specific characteristics of the input signal.

For spatial modeling, the channel correlation matrix \( H \) is generated using a  channel-dependent encoder  \( \mathcal{K}_\theta \), which processes the input patch \( \hat{x}^{(p)} \) and outputs channel-wise weights as shown in Figure~\ref{fig: illustration}. This encoder is implemented as a transformer block with self-attention mechanisms, ensuring that inter-channel dependencies are effectively captured.

Finally, to optimize the temporal-spatial combination, we enforce a reconstruction objective that encourages the model to approximate the original EEG patch by combining the temporal embeddings \( Z \) and spatial weights \( H \). The reconstruction loss is defined as:
\begin{equation}
\mathcal{L}_{\text{dec}} = \frac{1}{n} \sum_{i=1}^{n} \left\| \mathcal{N}(H, Z) - \hat{x}^{(p)} \right\|^2,
\end{equation}
where \( \mathcal{N}(\cdot, \cdot) \) denotes the matrix multiplication between spatial and temporal factors. This objective ensures that the temporal patterns and spatial weights jointly reconstruct the input signal, promoting better decomposition and alignment under cross-domain variations.

\begin{figure}[]
  \centering
  \includegraphics[width=\columnwidth, trim=0 9 0 0, clip]{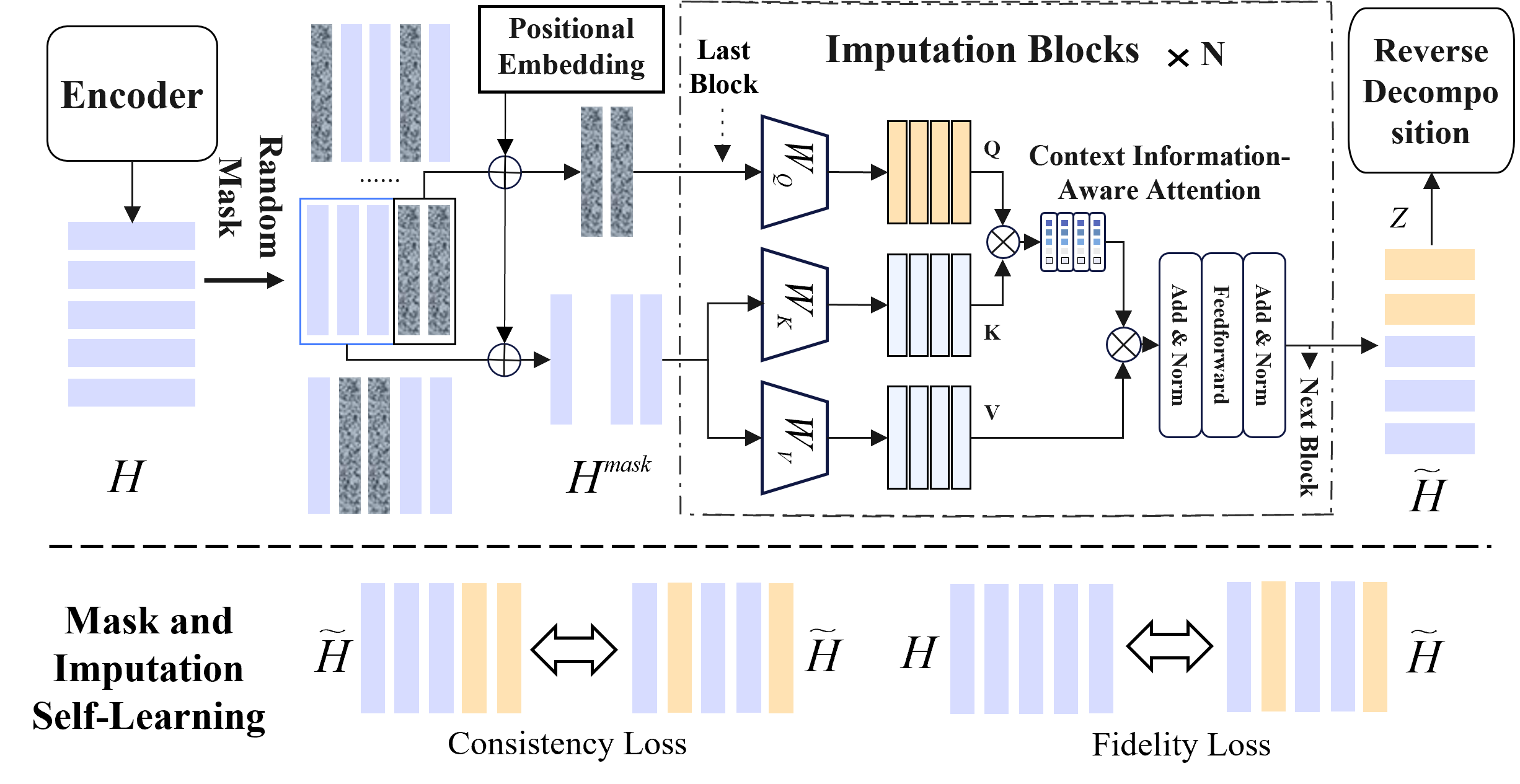}
  \caption{The architecture of the Channel-dependent Mask and Imputation Module.}
  \label{fig:CMIM}
\end{figure}

\subsection{Channel-dependent Mask and Imputation Module}

To address the spatial heterogeneity of EEG signals across different domains, there are prior works suggest that directly modifying or correcting the data can be more effective than merely mitigating domain gaps through feature transformation \cite{10.1093/nsr/nwaf086}. In line with this, we propose the  Channel Masking and Imputation Module. In contrast to existing approaches that focus on low-resolution upsampling \cite{wang2025generative}, CMIM introduces a channel-based masking mechanism and a context-aware imputation strategy to explicitly simulate and recover from cross-domain variability in EEG spatial patterns, as illustrated in Figure~\ref{fig:CMIM}.

Specifically, CMIM first applies random channel masking to the latent spatial representation \( H \in \mathbb{R}^{T' \times d} \), producing a masked version \( H^{mask} \), where certain channels (i.e., rows) are replaced by a  shared learnable mask token  sampled from a normal distribution:
\begin{equation}
H^{mask} = (H \odot M) + M_{\text{normal}},
\end{equation}
where \( M \in \{0,1\}^{T' \times d} \) is a binary mask matrix indicating masked positions, and \( \odot \) denotes element-wise multiplication. \( M_{\text{normal}} \) represents the randomly sampled placeholder token shared across masked entries.

Next,  positional embeddings  are added to \( H^{mask} \) to encode the spatial ordering of channels. The resulting representation is passed through a stack of  imputation blocks , each consisting of a  Context Information-Aware Attention module , layer normalization, and feed-forward layers. The attention mechanism is formulated as:
\begin{equation}
\text{Attention}(H^{mask}, H, H) = \text{softmax}\left(\frac{H^{mask} W_Q (H W_K)^\top}{\sqrt{d_k}}\right) H W_V,
\end{equation}
where \( W_Q, W_K, W_V \in \mathbb{R}^{d \times d_k} \) are learnable projection matrices, and \( d_k \) is the attention dimension. Here, masked tokens \( H^{mask} \) serve as  queries , while the full latent representation \( H \) provides the  keys  and  values , allowing reconstruction of missing channels based on global context. 

Our architecture employs a multi-layer imputation block design where, starting from the second layer, each block takes the previous layer's output as queries (q) while maintaining consistent keys (k) and values (v) from the original context information throughout all layers. This design preserves the integrity of the contextual information while allowing progressive refinement of the imputed features.
The final imputed representations \(\tilde{H}\) are obtained as the output of the last layer. For each masked region indexed by \(m(i)\), the imputer \(g_\phi\) generates the reconstructed embedding through:
\begin{equation}
\tilde{H}(i) = g_\phi\left(\text{Attn}_{\text{final}}(q), m(i)\right),
\end{equation}
where \(\text{Attn}_{\text{final}}(q)\) denotes the output from the final attention layer's query processing, and \(m(i)\) encodes the spatial position of the i-th missing segment.

To ensure high-quality imputation and improve generalization across domains, we introduce two complementary learning objectives:

\begin{itemize}
    \item \textbf{Fidelity Loss} ($\mathcal{L}_{\text{fid}}$): Encourages the imputed tokens $\tilde{H}(i)$ to be close to the ground-truth values $H(i)$, promoting accurate recovery:
    \begin{equation}
        \mathcal{L}_{\text{fid}} = \frac{1}{M} \sum_{i=1}^{M} \| \tilde{H}(i) - H(i) \|^2.
    \end{equation}
    
    \item \textbf{Consistency Loss} ($\mathcal{L}_{\text{cons}}$): Measures the discrepancy between imputed results under different random masks applied to the same sample. Let $H_n^m$ and $H_{n'}$ denote different masked versions of the same EEG segment; then:
    \begin{equation}
        \mathcal{L}_{\text{cons}} = \frac{1}{N} \sum_{n=1}^{N} \| g_\phi(H_n^m) - g_\phi(H_{n'}) \|^2.
    \end{equation}
\end{itemize}

The total loss combines both terms with a weighting factor $\lambda$:
\begin{equation}
    \mathcal{L} = \mathcal{L}_{\text{fid}} + \lambda \cdot \mathcal{L}_{\text{cons}}.
\end{equation}

This dual-objective setup ensures that the imputer produces  reliable  reconstructions while maintaining  consistency  across varying masking patterns, thereby enhancing robustness under distribution shifts.

Finally, to prevent noise from imputed channels from negatively impacting downstream classification, the imputation process is integrated as an auxiliary module into the overall IMAC architecture. A classification head \( \mathcal{C}_\theta \), implemented using EEGNet \cite{lawhern2018eegnet}, takes the full reconstructed signal as input and is jointly trained to optimize classification accuracy alongside the imputation loss. This encourages the model to balance between  effective feature recovery  and  discriminative learning , resulting in better cross-domain generalization.

\begin{figure}[]
  \centering
  \includegraphics[width=\columnwidth]{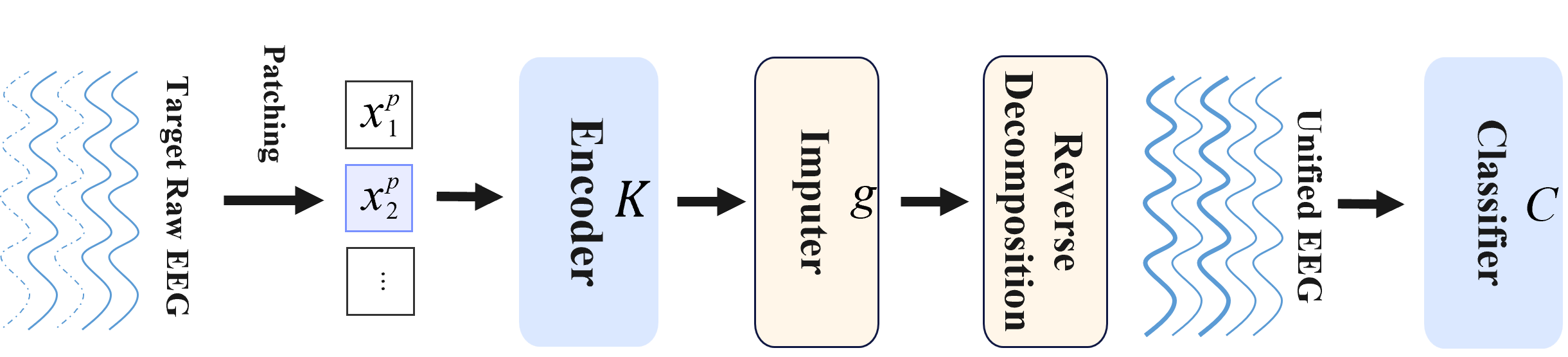}
  \caption{The inference process of the proposed IMAC.}
  \label{fig:inference}
\end{figure}

\subsection{Model Inference}
 
During inference on target EEG samples as shown in Figure~\ref{fig:inference}, IMAC adopts a deterministic pipeline that omits Spatial Topology Unification and random masking. The process begins by partitioning the raw multichannel EEG signals into spatial-temporal segments $\{x^p_1, x^p_2, \ldots\}$, preserving localized signal characteristics. Each segment is then processed through the pretrained encoder $K$ to extract spatial features, followed by the imputer $g_\phi$, which deterministically reconstructs missing channels using learned attention patterns.  

Subsequently, a reverse decomposition step transforms the features back into signal representations by multiplying them with temporal factors. These domain-invariant features are then fed into the frozen classifier $\mathcal{C}_\theta$ to generate the final prediction. By eliminating training-time stochasticity, this streamlined pipeline ensures robust handling of unseen channel configurations through stable, learned reconstruction patterns, while maintaining computational efficiency—a critical requirement for real-world deployment.

\section{Experiments}

\begin{table*}[htbp]
\centering
\caption{Cross-Center Results on 4 PD Recognition Datasets and 5 MI Classification Datasets, where each column represents the performance when the corresponding dataset is used as the test set.  
All values in the table represent Accuracy (\%).}
\resizebox{\textwidth}{!}{
\begin{tabular}{lc|cccc|c|ccccc|c}
\toprule
\multirow{2}{*}{\textbf{Method}} & \multirow{2}{*}{\textbf{Ref}} & \multicolumn{5}{c|}{\textbf{PD Task (Leave-One-Dataset-Out)}} & \multicolumn{6}{c}{\textbf{MI Task (Leave-One-Dataset-Out)}} \\
                &              & \textbf{UC} & \textbf{UNM} & \textbf{Iowa} & \textbf{Finland} & \textbf{Avg.} & \textbf{Cho} & \textbf{Physio} & \textbf{BNCI} & \textbf{Lee} & \textbf{Weibo} & \textbf{Avg.} \\
\midrule
EEGNet & TNSRE 2023 & 80.57 & 78.18 & 80.69 & 79.84 & 79.82 & 83.37 & 83.68 & 83.59 & 82.85 & 81.70 & 82.96 \\
EEGConformer & TNSRE 2023 & 80.43 & 79.15 & 80.92 & 81.02 & 80.38 & 84.26 & 85.33 & 84.50 & 85.11 & 86.75 & 85.19 \\
EEGDeformer & JHBI 2024 & 81.89 & 80.26 & 81.49 & 82.11 & 81.44 & 85.73 & 86.86 & 85.95 & 86.51 & 87.04 & 86.42 \\
MAPU & KDD 2023 & 81.12 & 80.03 & 80.84 & 82.15 & 81.04 & 85.12 & 84.43 & 87.24 & 86.11 & 86.95 & 85.97 \\

Medformer & NIPS 2024 & 82.12 & 79.53 & 81.04 & 82.35 & 81.26 & 85.52 & 85.93 & 85.44 & 86.25 & 86.18 & 85.95 \\
MCPNet & TIM 2024 & 79.90 & 81.21 & 82.01 & 80.33 & 80.87 & 85.39 & 87.61 & 87.42 & 85.02 & 85.98 & 86.48 \\

POND & KDD 2024 & 81.08 & 80.79 & 82.01 & 82.51 & 81.60 & \underline{88.48} & 85.79 & 85.68 & 85.91 & \underline{87.81} & 86.73 \\
EEGPT & NeurIPS 2024 & 79.58 & 80.61 & 79.89 & 80.34 & 80.11 & 84.22 & 84.51 & 84.20 & 81.45 & 84.35 & 84.15 \\
MASER & TNSRE 2024 & 80.26 & 80.57 & 80.38 & \underline{83.79} & 81.25 & 87.26 & \underline{88.57} & \underline{86.38} & \underline{87.69} & 87.50 & \underline{87.48} \\
SaSDim & IJCAI 2024 & \underline{82.19} & \underline{82.20} & \underline{82.01} & 83.32 & \underline{82.41} & 87.99 & 87.30 & 86.11 & 85.42 & 85.32 & 86.29 \\
\midrule
Ours &  & \textbf{83.18} & \textbf{84.77} & \textbf{84.91} & \textbf{84.96} & \textbf{84.46} & \textbf{90.23} & \textbf{90.78} & \textbf{91.14} & \textbf{90.96} & \textbf{90.79} & \textbf{90.92} \\
\bottomrule
\end{tabular}
}
\label{tab:cross-center}
\end{table*}

\begin{table*}[htbp]
\centering
\caption{Cross-Center Results on PD Recognition and MI Classification Tasks.  
The Leave-One-Dataset-Out setting is applied, where each dataset serves as the test set, with the remaining datasets used for training.}
\resizebox{\textwidth}{!}{
\begin{tabular}{l|ccccc|ccccc}
\toprule
\multirow{2}{*}{\textbf{Method}} & \multicolumn{5}{c|}{\textbf{PD Task  (Leave-One-Dataset-Out)}} & \multicolumn{5}{c}{\textbf{MI Task (Leave-One-Dataset-Out)}} \\
& \textbf{Accuracy} & \textbf{F1-score} & \textbf{Precision} & \textbf{Recall} & \textbf{Kappa} & \textbf{Accuracy} & \textbf{F1-score} & \textbf{Precision} & \textbf{Recall} & \textbf{Kappa} \\
\midrule
EEGNet & $79.82_{\scriptsize{\pm1.03}}$ & $79.43_{\scriptsize{\pm0.98}}$ & $79.93_{\scriptsize{\pm1.05}}$ & $79.14_{\scriptsize{\pm0.97}}$ & $63.44_{\scriptsize{\pm0.79}}$ & $82.96_{\scriptsize{\pm0.73}}$ & $82.63_{\scriptsize{\pm0.70}}$ & $83.12_{\scriptsize{\pm0.75}}$ & $82.27_{\scriptsize{\pm0.69}}$ & $66.03_{\scriptsize{\pm0.71}}$ \\
EEGConformer & $80.38_{\scriptsize{\pm1.03}}$ & $80.32_{\scriptsize{\pm1.49}}$ & $79.26_{\scriptsize{\pm1.53}}$ & $80.21_{\scriptsize{\pm1.29}}$ & $65.22_{\scriptsize{\pm0.93}}$ & $84.26_{\scriptsize{\pm1.44}}$ & $83.47_{\scriptsize{\pm1.28}}$ & $84.35_{\scriptsize{\pm1.12}}$ & $83.25_{\scriptsize{\pm1.39}}$ & $67.11_{\scriptsize{\pm0.98}}$ \\
EEGDeformer & $81.44_{\scriptsize{\pm1.02}}$ & $81.84_{\scriptsize{\pm0.98}}$ & $81.78_{\scriptsize{\pm1.12}}$ & $81.57_{\scriptsize{\pm0.79}}$ & $66.77_{\scriptsize{\pm0.84}}$ & $86.42_{\scriptsize{\pm0.90}}$ & $86.34_{\scriptsize{\pm0.98}}$ & $85.99_{\scriptsize{\pm1.42}}$ & $86.85_{\scriptsize{\pm1.44}}$ & $68.49_{\scriptsize{\pm0.87}}$ \\
MAPU & $81.04_{\scriptsize{\pm0.93}}$ & $80.83_{\scriptsize{\pm0.89}}$ & $81.24_{\scriptsize{\pm0.95}}$ & $80.43_{\scriptsize{\pm0.88}}$ & $64.64_{\scriptsize{\pm0.71}}$ & $85.97_{\scriptsize{\pm1.10}}$ & $85.13_{\scriptsize{\pm1.05}}$ & $85.63_{\scriptsize{\pm1.12}}$ & $84.73_{\scriptsize{\pm1.04}}$ & $68.05_{\scriptsize{\pm0.88}}$ \\
Medformer & $81.26_{\scriptsize{\pm1.20}}$ & $80.97_{\scriptsize{\pm1.15}}$ & $81.53_{\scriptsize{\pm1.22}}$ & $80.63_{\scriptsize{\pm1.14}}$ & $65.01_{\scriptsize{\pm0.92}}$ & $85.95_{\scriptsize{\pm0.55}}$ & $85.63_{\scriptsize{\pm0.53}}$ & $86.17_{\scriptsize{\pm0.56}}$ & $85.23_{\scriptsize{\pm0.52}}$ & $68.63_{\scriptsize{\pm0.44}}$ \\
MCPNet & $80.87_{\scriptsize{\pm0.87}}$ & $80.91_{\scriptsize{\pm0.83}}$ & $81.35_{\scriptsize{\pm0.89}}$ & $80.53_{\scriptsize{\pm0.82}}$ & $64.82_{\scriptsize{\pm0.67}}$ & $86.48_{\scriptsize{\pm1.14}}$ & $85.34_{\scriptsize{\pm1.09}}$ & $85.73_{\scriptsize{\pm1.15}}$ & $85.02_{\scriptsize{\pm1.08}}$ & $68.24_{\scriptsize{\pm0.91}}$ \\
POND & $\underline{81.60}_{\scriptsize{\pm0.76}}$ & $81.03_{\scriptsize{\pm0.73}}$ & $80.47_{\scriptsize{\pm0.78}}$ & $80.93_{\scriptsize{\pm0.72}}$ & $65.12_{\scriptsize{\pm0.58}}$ & $86.73_{\scriptsize{\pm1.29}}$ & $86.03_{\scriptsize{\pm1.23}}$ & $86.53_{\scriptsize{\pm1.31}}$ & $85.63_{\scriptsize{\pm1.22}}$ & $68.94_{\scriptsize{\pm1.03}}$ \\
EEGPT & $80.11_{\scriptsize{\pm0.47}}$ & $79.83_{\scriptsize{\pm0.45}}$ & $80.42_{\scriptsize{\pm0.48}}$ & $79.31_{\scriptsize{\pm0.44}}$ & $63.84_{\scriptsize{\pm0.36}}$ & $83.75_{\scriptsize{\pm1.31}}$ & $83.83_{\scriptsize{\pm1.25}}$ & $84.31_{\scriptsize{\pm1.33}}$ & $83.24_{\scriptsize{\pm1.24}}$ & $67.32_{\scriptsize{\pm1.05}}$ \\
MASER & $81.25_{\scriptsize{\pm1.53}}$ & $\underline{81.02}_{\scriptsize{\pm1.46}}$ & $\underline{81.53}_{\scriptsize{\pm1.55}}$ & $\underline{80.83}_{\scriptsize{\pm1.45}}$ & $\underline{65.04}_{\scriptsize{\pm1.17}}$ & $\underline{87.48}_{\scriptsize{\pm0.63}}$ & $\underline{87.74}_{\scriptsize{\pm0.60}}$ & $\underline{88.13}_{\scriptsize{\pm0.64}}$ & $\underline{87.32}_{\scriptsize{\pm0.59}}$ & $\underline{70.04}_{\scriptsize{\pm0.50}}$ \\
SaSDim & $82.43_{\scriptsize{\pm0.49}}$ & $82.14_{\scriptsize{\pm0.47}}$ & $82.53_{\scriptsize{\pm0.50}}$ & $81.73_{\scriptsize{\pm0.46}}$ & $66.51_{\scriptsize{\pm0.38}}$ & $86.43_{\scriptsize{\pm1.16}}$ & $86.03_{\scriptsize{\pm1.11}}$ & $86.53_{\scriptsize{\pm1.18}}$ & $85.53_{\scriptsize{\pm1.10}}$ & $69.03_{\scriptsize{\pm0.93}}$ \\
\midrule
Ours & $\mathbf{84.46}_{\scriptsize{\pm0.76}}$ & $\mathbf{84.13}_{\scriptsize{\pm0.73}}$ & $\mathbf{84.83}_{\scriptsize{\pm0.77}}$ & $\mathbf{83.73}_{\scriptsize{\pm0.72}}$ & $\mathbf{68.82}_{\scriptsize{\pm0.58}}$ & $\mathbf{90.78}_{\scriptsize{\pm0.35}}$ & $\mathbf{90.63}_{\scriptsize{\pm0.34}}$ & $\mathbf{91.03}_{\scriptsize{\pm0.36}}$ & $\mathbf{90.23}_{\scriptsize{\pm0.33}}$ & $\mathbf{72.83}_{\scriptsize{\pm0.28}}$ \\
\bottomrule
\end{tabular}
}
\label{tab:average}
\end{table*}

\begin{table}[htbp]
\centering
\caption{Cross-Subject Results on emotion recognition, PD recognition and MI classification task. The Leave-One-Subject-Out setting is applied, where each subject as the test set, with the remaining subjects used for training.}
\resizebox{0.42\textwidth}{!}{
\begin{tabular}{l|ccc|c}
\toprule
\textbf{Method} & \textbf{DEAP} & \textbf{UC} & \textbf{Cho} & \textbf{Avg.} \\
\midrule
EEGNet  & 69.76 & 87.88 & 91.02 & 82.22 \\
EEGConformer & 70.59 & 86.58 & 90.81 & 82.66 \\
EEGDeformer & 71.39 & 87.65 & 91.34 & 83.46 \\
Medformer & 70.13 & 86.09 & 91.76 & 82.66 \\
MCPNet  & 71.45 & 87.92 & 90.83 & 83.07 \\
MAPU  & 71.56 & 87.99 & 92.05 & 83.53 \\
POND  & \underline{72.23} & \underline{88.02} & \underline{92.13} & \underline{84.13} \\
\midrule
Ours & \textbf{73.12} & \textbf{89.11} & \textbf{94.16} & \textbf{85.13} \\
\bottomrule
\end{tabular}
\label{tab:cross-subject}
}
\end{table}

\subsection{Datasets} 

In this study, we evaluate IMAC on 10 public EEG datasets covering Parkinson’s Disease (PD), motor imagery (MI), and emotion recognition tasks: PD datasets include \textbf{UC} ($41$ ch, $35$ subj) \cite{jackson2019characteristics}, \textbf{UNM} ($64$ ch, $42$ subj) and \textbf{Iowa} ($50$ ch, $37$ subj) \cite{anjum2020linear}, and \textbf{Finland} ($41$ ch, $40$ subj) \cite{railo2020deficits}. \textbf{MI} datasets include \textbf{Cho2017} ($32$ ch, $52$ subj) \cite{cho2017eeg}, \textbf{PhysionetMI} ($64$ ch, $109$ subj) \cite{schalk2004bci2000}, \textbf{BNCI2014001} ($22$ ch, $10$ subj) \cite{tangermann2012review}, \textbf{Lee2019MI} ($62$ ch, $54$ subj) \cite{lee2019eeg}, and \textbf{Weibo2014} ($30$ ch, $10$ subj) \cite{yi2014evaluation}. Emotion recognition is evaluated on \textbf{DEAP} ($32$ ch, $32$ subj) \cite{5871728}.

\subsection{Baselines and Implementation Details} 

To comprehensively evaluate and compare our method with relevant baseline approaches, we selecte a range of state-of-the-art models. Among these, EEGConformer \cite{9991178}, EEGDeformer \cite{10763464}, Medformer \cite{wang2024medformer}, and EEGNet \cite{lawhern2018eegnet} serve as models specifically designed for EEG classification tasks. MCPNet \cite{qiu2024novel}, MAPU \cite{ragab2023source}, and POND \cite{wang2023prompt} represent cutting-edge domain generalization methods tailored for time series classification. MASER \cite{10720236} and SaSDim \cite{ijcai2024p283} are advanced spatial time series imputation methods, leveraging unique architectures to enhance spatial resolution and dynamic temporal modeling. EEGPT \cite{NEURIPS2024_4540d267} is a $10$-million-parameter pretrained Transformer model capable of extracting effective features from low-resolution EEG signals across diverse domains.

We evaluate the models using accuracy as the primary metric, while also reporting F1-score, precision, recall, and Cohen's kappa as additional indicators under the cross-center experimental setting. To ensure a fair comparison, we reproduce all the methods and conducted experiments under both the Leave-One-Dataset-Out and Leave-One-Subject-Out settings.
Specifically, for the Leave-One-Dataset-Out setting in the Parkinson's Disease (PD) task, we select the same $32$ channels from the UC San Diego dataset\cite{qiu2024novel}, ensuring that each PD dataset consists of $32$ channels. We then select one dataset as the test set while using the remaining datasets for training. As shown in Table~\ref{tab:cross-center}, each column represents the EEG classification results when a specific dataset is used as the test set.
For the Motor Imaginary (MI) task, we select $22$ channels from the BCNI dataset, ensuring that each MI dataset contains $22$ channels.

Beyond cross-dataset evaluation, we also assessed IMAC under a Leave-One-Subject-Out (LOSO) protocol, in which each subject is held out in turn for testing \cite{wang2024dmmr}. The results for DEAP, UC, and Cho are shown in Table~\ref{tab:cross-subject}, where each column represents the average performance of all subjects in a particular dataset when used as the test set. In both tables, the "Avg." column indicates the average performance of a method across different datasets under the respective experimental setting.

\subsection{Results of Cross-domain EEG Classification}

Table~\ref{tab:cross-center} displays the accuracy of all methods on each dataset and Table~\ref{tab:average} shows the average classification performance in the two tasks under the Leave-One-Dataset-Out settings. Our proposed IMAC consistently outperforms all baselines in both tasks across the $9$ cross-center scenarios, demonstrating superior generalization ability under cross-domain data shifts. Specifically, 
IMAC achieves $84.46$\% accuracy on the PD recognition task and $90.78$\% on the MI classification task, representing improvements of $2.03$\% and $2.30$\% respectively over the second-best method, MASER. In terms of F1-score, precision, recall, and kappa, IMAC also ranks the highest across both tasks, underscoring its ability to maintain balanced and reliable classification performance.

The most notable performance gain is observed on the BNCI dataset, which is known for its missing channels and inconsistent electrode configurations. As shown in Table~\ref{tab:cross-center}, our IMAC model achieves an accuracy of $91.14$\% on the MI task in this challenging scenario, surpassing the second-best method (MASER: $86.38$\%) by $4.76$\%. This result suggests that IMAC’s channel-aware masking and imputation strategy is particularly effective in mitigating performance degradation caused by severe spatial distortions and non-uniform sensor setups.
Among the baselines, SaSDim and MASER show the strongest results overall. SaSDim achieves the second-best performance on the PD task with $82.43$\% accuracy, while MASER ranks second on the MI task with $87.48$\%. Nonetheless, these methods still trail behind IMAC across all metrics, particularly under high-variance and low-resolution conditions, which are exacerbated in cross-center evaluations.

Although large-scale brain models like EEGPT have demonstrated strong capabilities in extracting cross-domain EEG features, their performance under cross-center settings remains suboptimal. As shown in Table~\ref{tab:average}, EEGPT achieves $80.11$\% accuracy on the PD task and $83.75$\% on the MI task—significantly lower than the results achieved by our proposed IMAC framework.
In addition, the EEGNet classifier yields the poorest performance under real-world domain shifts, further underscoring the limitations of conventional models in handling complex distribution discrepancies. These findings highlight that strong feature extraction alone is insufficient; without effective spatio-temporal alignment at the signal level, such models struggle to cope with practical deployment challenges—particularly those involving severe variations in sensor locations and heterogeneous electrode configurations.

Table~\ref{tab:cross-subject} presents the results of the Leave-One-Subject-Out experiments for emotion recognition, PD recognition, and MI classification tasks. 
In comparison with the results in Table~\ref{tab:cross-center}, domain generalization methods like MCPNet, MAPU and POND demonstrate strong performance. Specifically, the second-performing method POND acieves $84.13$\%, indicating their effectiveness in capturing domain-invariant information when addressing temporal variability in cross-subject scenarios. 
While our proposed method maintains a consistent advantage across all three tasks, achieving an overall average accuracy of $85.13$\%, which is $1$\% higher than the second-best method, POND.
Moreover, on challenging tasks such as emotion recognition, our IMAC performs the best accuracy of $73.12$\%, excels by $0.89$\% than the second performing POND,  emphasizing the robustness of our method in handling domain shifts caused by subject-specific variations.

\begin{figure}[]
  \centering
  \includegraphics[width=0.95\columnwidth]{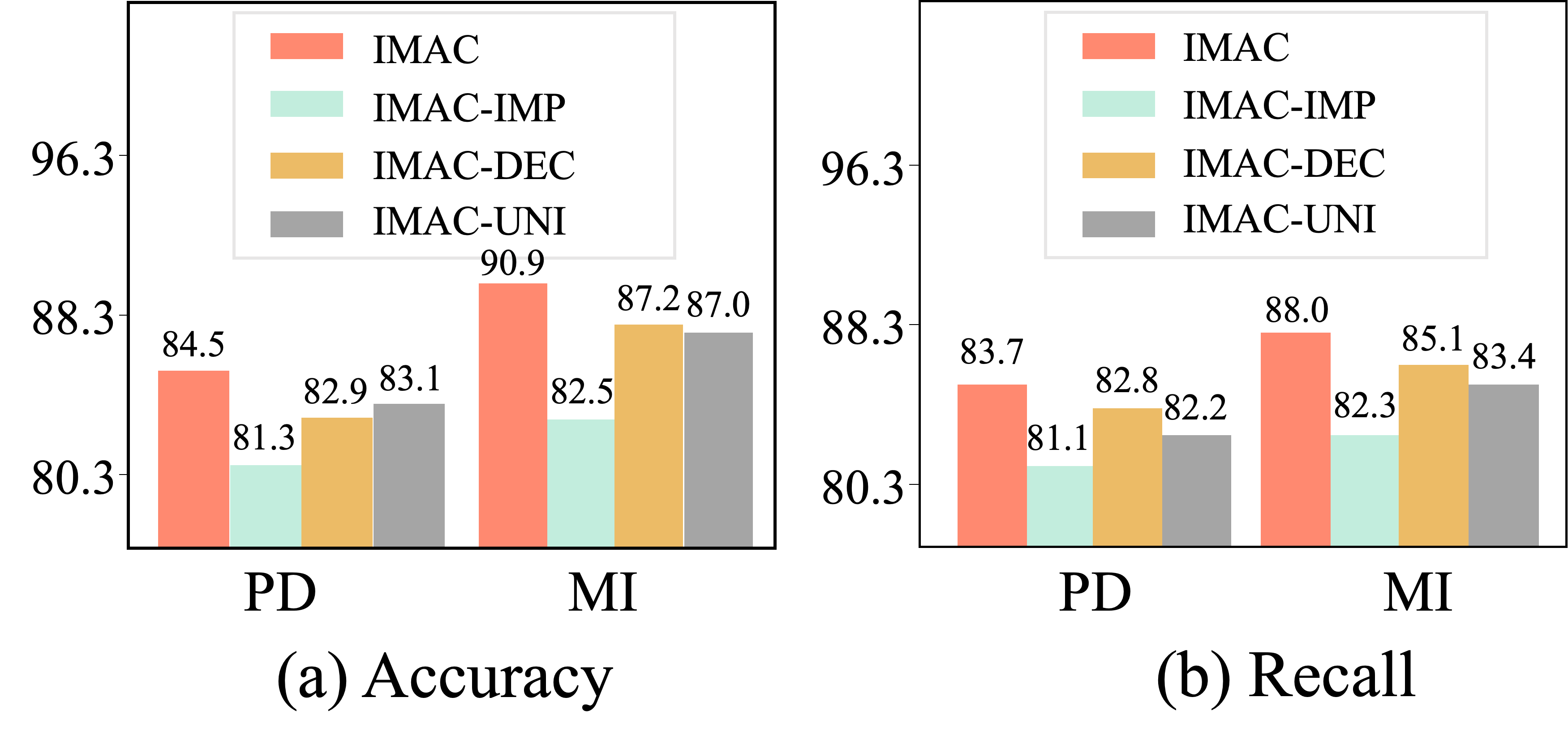}
  \caption{Ablation study performance comparison between IMAC and three variant models on two EEG classification tasks.}
  \label{fig:ablation}
\end{figure}

\begin{figure}[]
  \centering
  \includegraphics[width=0.95\columnwidth]{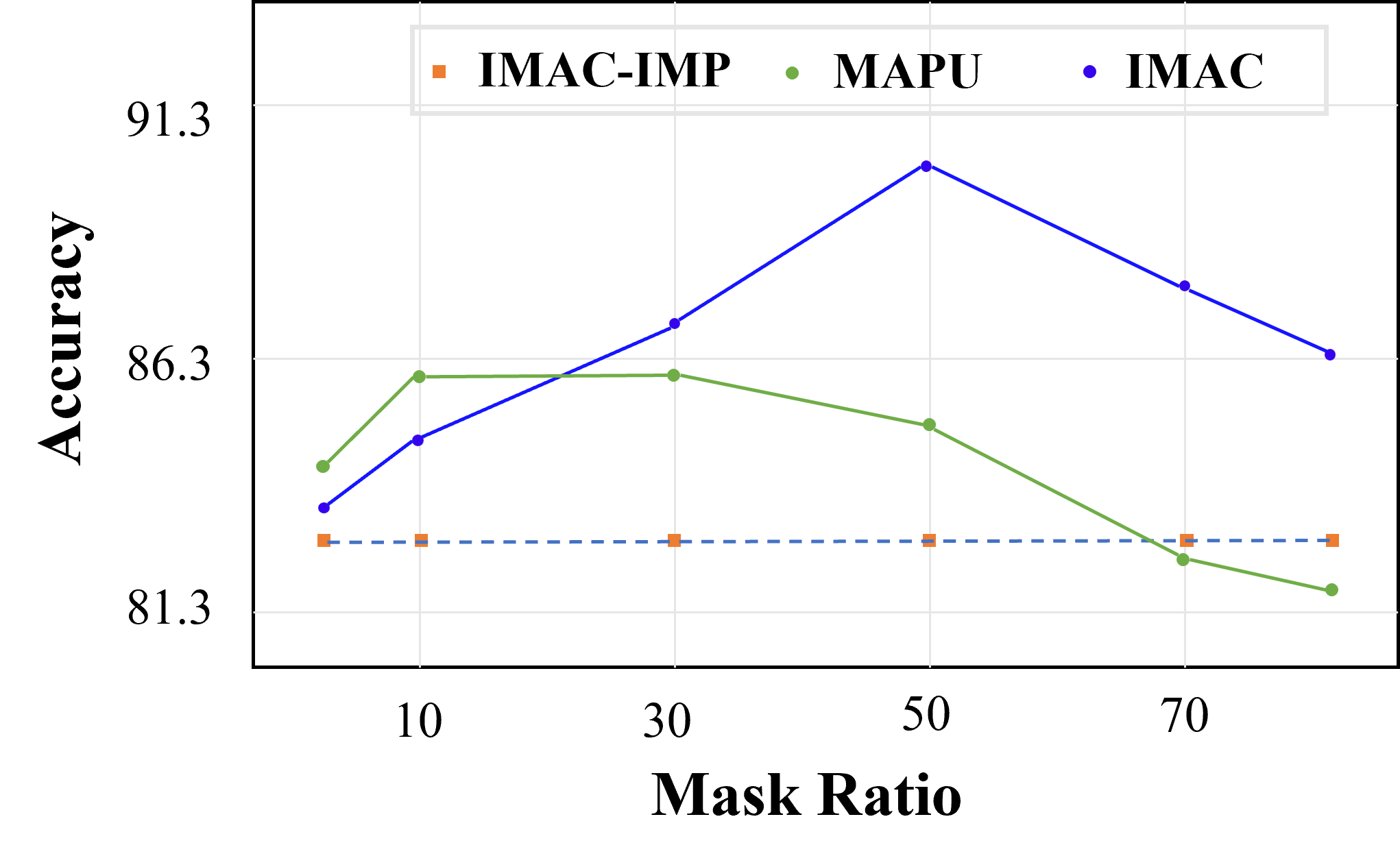}
  \caption{Analysis of adaptation performance with varying mask ratio for self-supervised learning methods. The blue, green and yellow points represent the accuracy results from IMAC, MAPU and IMAC-IMP, respectively.}
  \label{fig:mask}
\end{figure}

\subsection{Ablation Study}

To evaluate the contribution of each module in IMAC for addressing real-world data shift problems, we conducted an ablation study comparing IMAC with three variants. \textbf{IMAC-IMP} retains only the matrix factorization module. \textbf{IMAC-DEC} is a version of IMAC without the matrix factorization module. \textbf{IMAC-UNI} removes the spatial topology unification module. The challenging Leave-One-Dataset-Out setting is utilized to test the performance.

Figure~\ref{fig:ablation} presents the accuracy and recall performance of these variants compared to the full IMAC model. Overall, IMAC achieves the best performance when all three modules are present, underscoring their necessity for handling real-world data shifts effectively.  
Notably, the largest performance drop is observed in the MI task. IMAC achieves an accuracy of $90.9$\%, while IMAC-IMP only reaches $82.5$\%, resulting in a significant accuracy decline of $8.4$\%. This highlights the challenges posed by the inherent data shifts in MI tasks, which are characterized by substantial differences in EEG collection paradigms. Without the Channel-dependent Mask and Imputation Module, IMAC-IMP fails to capture a unified topology, leading to decreased classification accuracy.  

Among the variants, IMAC-IMP performs the worst across all scenarios, with a $3$\% to $8$\% drop in both accuracy and recall for both tasks. This is because it relies solely on spatial correlations between electrode locations, neglecting the spatial alignment of EEG signals.  
Both IMAC-DEC and IMAC-UNI exhibit better performance than IMAC-IMP but still lag behind the full IMAC model. IMAC-DEC applies the encoder $\mathcal{K}_\theta$ directly for temporal-spatial feature extraction, while IMAC-UNI operates on EEG signals with fewer channels, omitting spatial topology unification. These variants show a $2$\% reduction in accuracy and recall in the PD task and a $4$\% reduction in the MI task compared to IMAC.  

The masking ratio during self-supervised pretraining is crucial, as
it defines the balance between task difficulty and representation
quality. Further, to explore how IMAC performs under varying levels of missing data, we evaluate its accuracy at six specific masking ratios: $5$\%, $10$\%, $30$\%, $50$\%, $70$\%, and $80$\%. As shown in Figure~\ref{fig:mask}, IMAC's performance improves as the mask ratio increases, peaking at 50\%, before declining at higher ratios. This indicates that moderate masking enables IMAC to better exploit its cross-channel imputation mechanism, while excessive masking reduces available context and hampers reconstruction quality. Notably, even at extreme ratios such as $80$\%, IMAC still outperforms the baseline without imputation, demonstrating its robustness.

To further assess imputation strategies under varying degrees of missing information, we include MAPU as a comparison method. MAPU exhibits a contrasting trend: its performance remains relatively stable across different mask ratios, with only slight fluctuations. This suggests that MAPU's design may limit its sensitivity to increased masking, potentially due to a lack of adaptive spatial modeling. Compared with IMAC, MAPU performs competitively at low mask ratios (e.g., $5$\%, $10$\%) but fails to scale effectively as the imputation challenge intensifies. The comparison underscores IMAC’s superior capacity for learning informative representations from partially observed EEG data, especially when large portions of the input are missing.


\begin{figure}[]
  \centering
  \includegraphics[width=\columnwidth]{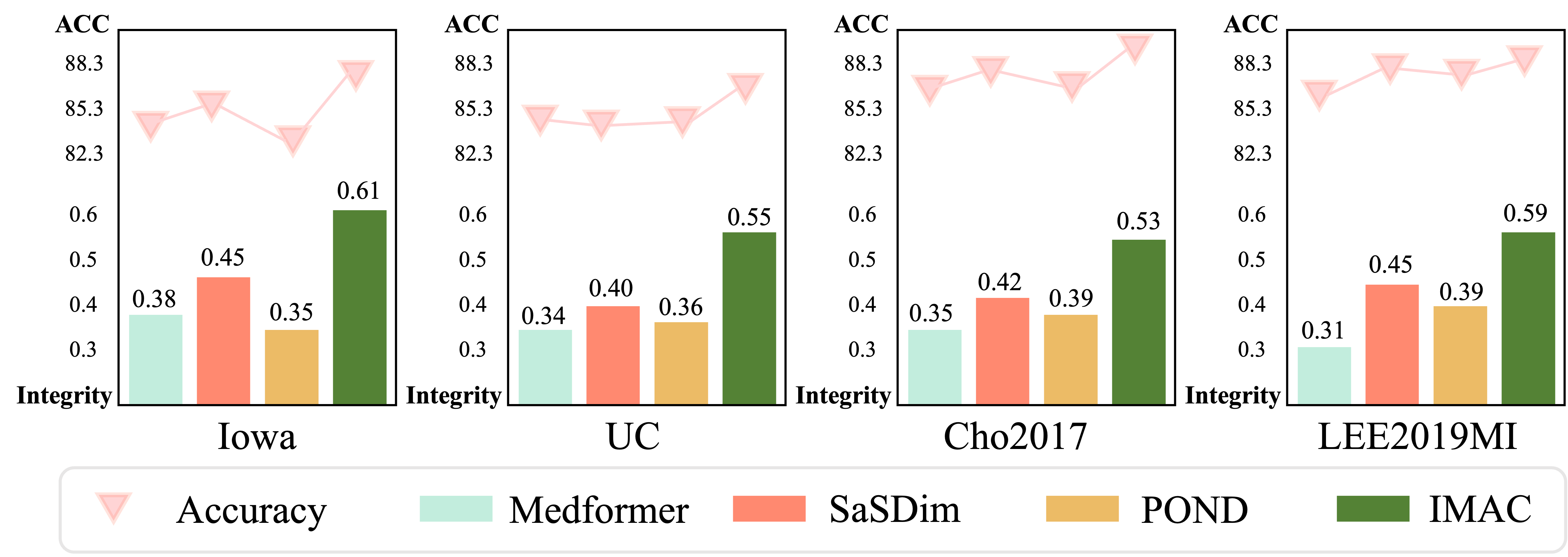}
  \caption{Analysis of simulated real-world data shifts on IMAC, Medformer, SaSDim, and POND across four datasets.}
  \label{fig:realworld}
\end{figure}

\subsection{Signal Integrity Measurement in Real-world Simulation }

Although our experiments on $10$ datasets already highlight the superior performance of IMAC under real-world data variability, we further evaluate its generalizability in simulated real-world data shifts during actual deployment. Following the study of \cite{wagh2022evaluating}, we simulate real-world data shifts by introducing noise and variability inherent in diverse datasets and emulate the data acquisition process across different devices by altering data characteristics. 
The following three types of data shifts were applied to different individuals within the same dataset:

\begin{itemize}
\item Application of a (1-25 Hz) band-pass filter;
\item Addition of broadband or narrowband noise ($\sigma=0.1$);
\item Implementation of a 50\% channel mask.
\end{itemize}

We measure the integrity of extracted features in the more challenging cross-center deployment using Delaunay neighborhood graphs, and compare classification results with well-performed Medformer, SaSDim, POND, and the proposed IMAC on the modified and unmodified versions of the LEE2019MI, UC, Cho2017, and Iowa datasets.
As shown in Figure~\ref{fig:realworld}, IMAC consistently outperforms the comparison methods in terms of integrity across all four datasets. Specifically, IMAC achieves the highest integrity scores on the modified versions of the Iowa, UC, Cho2017, and LEE2019MI datasets, demonstrating its robustness to real-world data shifts and noise. IMAC outperforms Medformer and POND by up to $35$\%$\sim$$74$\% in integrity, with the largest performance gap observed on the Iowa dataset, where IMAC achieves an integrity score of $0.61$, compared to SaSDim’s $0.45$ and Medformer’s $0.38$. This improvement highlights the effectiveness of IMAC in maintaining feature integrity under challenging real-world conditions.

\section{Conclusion}

In this paper, we introduce IMAC, a novel framework designed to tackle the challenges of cross-domain data shifts in EEG classification. By applying channel masking to simulate domain shifts, IMAC enables cross-domain knowledge transfer and learns to reconstruct consistent signal representations through a spatial imputation process. Extensive experiments on $10$ EEG classification datasets including emotion recognition, Parkinson’s disease detection, and motor imagery tasks, demonstrating that IMAC consistently outperforms state-of-the-art methods. The effectiveness of IMAC highlights the potential of incorporating spatial imputation to address a wide array of adaptation challenges.

\begin{acks}
This work was supported by the National Key Research and Development Program of China (Grant No. 2022ZD0118001), the National Natural Science Foundation of China (Grant Nos. U22A2022, 62332017), and was conducted in part at the MOE Key Laboratory of Advanced Materials and Devices for Post-Moore Chips, the Beijing Key Laboratory of Big Data Innovation and Application for Skeletal Health Medical Care, and the Beijing Advanced Innovation Center for Materials Genome Engineering.
\end{acks}

\bibliographystyle{ACM-Reference-Format}
\balance
\bibliography{acmart}

\end{document}